\PassOptionsToPackage{table}{xcolor}
\pdfoutput=1
\documentclass[11pt,a4paper,xcolor=table]{article}
\usepackage[hyperref]{acl2021}
\usepackage{times}
\usepackage{latexsym}
\usepackage[table]{xcolor}
\usepackage{graphicx}

\usepackage{amsmath}
\usepackage{booktabs}
\usepackage{multirow}
\usepackage{enumitem}
\usepackage{fdsymbol}
\usepackage{float}
\usepackage{comment}
\PassOptionsToPackage{hyphens}{url}\usepackage{hyperref}
\usepackage{xurl}

\usepackage{microtype}

\definecolor{stringcolor}{HTML}{FFF2CC}

\title{The NCTE Transcripts: \\ A Dataset of Elementary Math Classroom Transcripts}

  \author{Dorottya Demszky \\
  Stanford University  \\
  \texttt{ddemszky@stanford.edu} \\\And
  Heather Hill \\
  Harvard University \\
  \texttt{heather\_hill@gse.harvard.edu} \\}

\begin{document}
\maketitle
\begin{abstract}
Classroom discourse is a core medium of instruction – analyzing it can provide a window into teaching and learning as well as driving the development of new tools for improving instruction. We introduce the largest dataset of mathematics classroom transcripts available to researchers, and demonstrate how this data can help improve instruction. The dataset consists of 1,660 45-60 minute long 4th and 5th grade elementary mathematics observations collected by the National Center for Teacher Effectiveness (NCTE) between 2010-2013. The anonymized transcripts represent data from 317 teachers across 4 school districts that serve largely historically marginalized students. The transcripts come with rich metadata, including turn-level annotations for dialogic discourse moves, classroom observation scores, demographic information, survey responses and student test scores. We demonstrate that our natural language processing model, trained on our turn-level annotations, can learn to identify dialogic discourse moves and these moves are correlated with better classroom observation scores and learning outcomes. This dataset opens up several possibilities for researchers, educators and policymakers to learn about and improve K-12 instruction. The dataset can be found at \url{https://github.com/ddemszky/classroom-transcript-analysis}.
\end{abstract}

\section{Introduction}
\label{sec:intro}

Improving K-12 mathematics instruction in the aftermath of the Covid-12 pandemic is a major national priority, drawing support from both the U.S. government (U.S. Department of Education)
\footnote{\url{https://www.ed.gov/news/press-releases/us-department-education-announces-over-220-million-dollars-investments-government-private-and-public-sectors-support-student-recovery}})
and major foundations (e.g., Gates, Spencer).
\footnote{\url{https://www.spencer.org/news/announcing-covid-19-related-special-grant-cycle}}
A key step in this direction is to measure and facilitate the use of effective mathematics teaching practices, an effort that draws on a long history of research \citep[e.g.,][]{brophy1984teacher,sedova2019those}. Instructional measurement has traditionally relied on resource-intensive classroom observation. Recent natural language processing models, trained on manually scored classroom transcripts, enable measuring effective instructional practices in scalable and adaptable ways \citep{kelly2018automatically,suresh2019automating,demszky2021measuring,alic2022computationally,hunkins2022beautiful}. However, a common barrier to evaluating such measures is the lack of comprehensive data sources that link classroom transcripts to external variables, such as student and teacher demographics and learning outcomes.

To address this, we introduce a dataset of 1,660 U.S. 4th and 5th grade elementary math classroom transcripts collected by the National Center for Teacher Effectiveness (NCTE) between 2010-2013 \citep{kane2015national}. The anonymized transcripts represent data from 317 teachers across 4 school districts  serving largely historically marginalized student populations. The transcripts are associated with a wide range of metadata: (i) turn-level annotations for discourse moves, (ii) classroom observation scores, (iii)  questionnaires that capture teacher background, beliefs and classroom practices, (iv) student administrative data, (v)  questionnaires describing student background and classroom experiences, (vi) value added scores, which estimate teachers' contribution to students' academic performance. To our knowledge, this is the largest dataset of math classroom transcripts with linked outcomes available to researchers.

To illustrate how this data can be used to identify effective instructional practices, we build classifiers for discourse moves and validate these measures by correlating them with instructional outcomes. These discourse moves include on vs off task instruction, teachers' uptake of student ideas \citep{demszky2021measuring}, teachers' focusing questions \citep{alic2022computationally} and student reasoning --- the latter three of which are indicators of dialogic instruction, where students are active participants of the learning process \citep{bakhtin1986dialogic,nystrand1997opening,wells1999dialogic,alexander2008dialogic}.

We show that a RoBERTa classifier can learn to predict these discourse moves with moderate to high accuracy, leaving some room for improvement for future work. Importantly, we find that predictions for all of these discourse moves correlate significantly with classroom observation scores that measure instructional quality, teacher sensitivity, and classroom climate, among other items, while controlling for teacher and classroom covariates. Predictions for dialogic moves (teacher uptake, focusing questions and student reasoning) also show a significant positive correlation with teachers' value added scores. Taken together, these results demonstrate the value of this dataset for developing measures that can help us understand and facilitate effective instruction, for example, by powering automated feedback tools for teachers \citep{suresh2021using,demszky2021can}.

\section{Related Work}

We provide an overview of related corpora and methods pertaining to the computational analysis of classroom discourse.

\subsection{Related Corpora}
There is a wide range conversational datatasets available to researchers, capturing phone conversations \citep{godfrey1997switchboard}, task oriented dialogue \citep{budzianowski2018multiwoz}, meeting transcripts \citep{gapcorpus18}, among many others. These datasets can provide valuable insights about social dynamics in conversations, but to understand teaching and learning, we need to capitalize on datasets from the educational domain.

Conversational datasets in the education domain are scarce, due to resource intensiveness of data collection and privacy protections. The Measures of Effective Teaching (MET) dataset \citep{kane2013have} is the most similar to the NCTE data in terms of availability of outcomes. The MET data contains ~2,500 4-9th grade classroom recordings collected between 2009-2011 in six U.S. school districts. The recordings cover a variety of subjects, including English Language Arts and mathematics, and the recordings come with classroom observation scores, teacher and student demographic data and teacher value added scores. Although this dataset is a rich resource for studying instruction, it is challenging to work with, as it requires a paid subscription and interfacing with the data via a remote server that does not currently support many types of machine learning analyses. Furthermore, the MET data does not include transcripts; although subsets of the data have been transcribed as part of various research projects \citep{liu2020measuring,hunkins2022beautiful}, these transcripts are only accessible to these respective research teams.

The TalkMoves dataset \citep{suresh2022talkmoves} is currently the largest publicly available collection of transcripts o U.S. math instruction. The TalkMoves data contains 567 K-12 math classroom transcripts, annotated for talk moves based on accountable talk theory \citep{michaels2010accountable} and dialog acts from Switchboard \citep{Jurafsky:97-damsl}. The NCTE dataset complements the TalkMoves data in terms of availability of observation scores and outcomes. Other public datasets of educational interactions outside of the K-12 domain include the CIMA tutoring dataset \citep{stasaski-etal-2020-cima}, STEM lectures captured in the DRYAD dataset \citep{reimer2016evaluating} and the Coursera forum discussion dataset \citep{coursera-iri2014}, among others.

\subsection{Computational Analysis of Educational Interactions}

Our detection of discourse moves relates to a long-standing line of work on dialog act classification. The dialog act classification literature focuses on domain-agnostic dialog acts, e.g. acknowledgment, repetition, questioning, etc. In this work, we focus on detecting discourse moves that are indicators of better math instruction, similarly to the TalkMoves project \citep{suresh2019automating}. \citet{suresh2019automating} train transformer-based classifiers, including BERT \citep{devlin2019bert} and RoBERTa \citep{liu2019roberta}, on annotations for six accountable talk moves, such as keeping everyone together, getting students to relate, revoicing, pressing for reasoning. The NCTE data comes with annotations for related discourse moves (e.g. revoicing in TalkMoves and uptake in NCTE are similar constructs); predicting accountable talk moves on the NCTE data and studying correlations with outcomes is a promising direction for future work.

To better understand linguistic indicators of teacher effectiveness, \citet{liu2020measuring} analyze English Language Arts classes in the MET dataset \citep{kane2013have} using topic models, LIWC \citep{pennebaker2001linguistic} and an open-ended question classifier. The authors conduct factor analysis on several linguistic features and find that a factor indicating interactive, student centered instruction correlates positively with teachers' value added scores. \citet{hunkins2022beautiful} annotate transcripts of 156 video clips from 6-8th grade math classrooms in the MET dataset for teacher talk moves that support belonging and inclusivity, such as praise, admonishment, controlling language and learning mindset supportive language. They build a random forest classifier to predict these talk moves, and find that admonishment, for example, has a negative correlation with students' perception of the classroom environment. Finally, closely related work on the NCTE data has demonstrated the positive correlation between teacher uptake \citep{demszky2021measuring} and focusing questions \citep{alic2022computationally} and instructional outcomes. In this work, we expand our correlational analyses to all discourse moves annotated on the NCTE dataset.

\section{Dataset Description}
\label{sec:data}

\begin{table}[t!]
\centering
\renewcommand{\arraystretch}{1}
\begin{tabular}{lc}

\hline
\multicolumn{2}{c}{\cellcolor[HTML]{C3F8FF} \textbf{Transcripts}}                                                                                                                 \\ \hline
\cellcolor[HTML]{dffbff} \# of Transcripts                                                        & $1660$                                            \\
\cellcolor[HTML]{dffbff}Year 1                                                                            & $697$                                                      \\
\cellcolor[HTML]{dffbff}Year 2                                                                            & $616$                                                      \\
\cellcolor[HTML]{dffbff}Year 3                                                                            & $347$                                                     \\
\cellcolor[HTML]{dffbff}\begin{tabular}[c]{@{}l@{}}Avg \# Transcripts Per Teacher \end{tabular} & \begin{tabular}[c]{@{}c@{}}$5.24$\\ ($\pm2.46$)\end{tabular}    \\

\cellcolor[HTML]{dffbff}Avg \# of Turns                                                                   & \begin{tabular}[c]{@{}c@{}}$350$\\ ($\pm186$)\end{tabular}      \\
\cellcolor[HTML]{dffbff}Avg \% of Teacher Turns                                                           & \begin{tabular}[c]{@{}c@{}}$50.2\%$\\ ($\pm4.8\%$)\end{tabular} \\
\cellcolor[HTML]{dffbff}Avg \# of Words                                                                   & \begin{tabular}[c]{@{}c@{}}$5733$\\ ($\pm1782$)\end{tabular}    \\
\cellcolor[HTML]{dffbff}Avg \% of Teacher Words*                                                          & \begin{tabular}[c]{@{}c@{}}$87.7\%$\\ ($\pm7\%$)\end{tabular}   \\ \hline
\hline
\multicolumn{2}{c}{\cellcolor[HTML]{D4F1BA}\textbf{Teachers}}                                                                                                                 \\ \hline
\cellcolor[HTML]{EDF6E5}\# of Teachers                                                             & 317                                                             \\
\cellcolor[HTML]{EDF6E5}\% Male                                                                    & 16\%                                                            \\
\cellcolor[HTML]{EDF6E5}\% Black                                                        & 22\%                                                            \\
\cellcolor[HTML]{EDF6E5}\% Asian                                                                   & 3\%                                                             \\
\cellcolor[HTML]{EDF6E5}\% Hispanic/Latinx                                                         & 3\%                                                             \\
\cellcolor[HTML]{EDF6E5}\% White                                                                   & 65\%                                                            \\
\cellcolor[HTML]{EDF6E5}\begin{tabular}[c]{@{}l@{}}Avg \# of Years of Experience\end{tabular}    & \begin{tabular}[c]{@{}c@{}}10.23\\ (7.28)\end{tabular}          \\
\cellcolor[HTML]{EDF6E5}\begin{tabular}[c]{@{}l@{}}U.grad or Grad  Degree in Math\end{tabular} & 6\%                                                             \\
\cellcolor[HTML]{EDF6E5}BA in Education                                                            & 53\%                                                            \\
\cellcolor[HTML]{EDF6E5}Masters Degree                                                             & 76\%                                                            \\ \hline \hline
\multicolumn{2}{c}{\cellcolor[HTML]{FFEBAD}\textbf{Students}}                                                                                                                 \\ \hline
\cellcolor[HTML]{FFF6D9}\# of Students                                                             & 10,817                                                           \\
\cellcolor[HTML]{FFF6D9}Grade                                                                      & \begin{tabular}[c]{@{}c@{}}4th (51\%)\\ 5th (47\%)\end{tabular} \\
\cellcolor[HTML]{FFF6D9}\% Male                                                                    & 50\%                                                            \\
\cellcolor[HTML]{FFF6D9}\% African American                                                        & 43\%                                                            \\
\cellcolor[HTML]{FFF6D9}\% Asian                                                                   & 8\%                                                             \\
\cellcolor[HTML]{FFF6D9}\% Hispanic/Latinx                                                         & 23\%                                                            \\
\cellcolor[HTML]{FFF6D9}\% White                                                                   & 23\%                                                            \\
\cellcolor[HTML]{FFF6D9}\% Free or Reduced Lunch                                                   & 67\%                                                            \\
\cellcolor[HTML]{FFF6D9}\% Special Education Status                                                & 13\%                                                            \\
\cellcolor[HTML]{FFF6D9}Limited English Proficiency                                                & 21\%                                                            \\ \hline
\end{tabular}

\caption{Statistics on transcripts and on teachers and students who are mappable to these transcripts via administrative data. For each demographic, we use the naming convention from \citet{kane2015national}. Percentages for student grade levels do not add up to 100\% due to missing values. }
\label{tab:teacher_student_stats}
\end{table}

The dataset consists of 1,660 anonymized transcripts of whole lessons, collected as part of the National Center for Teacher Effectiveness (NCTE) Main Study \citep{kane2015national}.\footnote{Parents and teachers gave consent for the study (Harvard IRB \#17768),
and for de-identified data to be publicly shared for research.} The observations took place between 2010-2013 in 4th and 5th grade elementary math classrooms across four districts serving largely historically marginalized students. In the first two years, teachers and students were assigned to each classroom according to their school's usual procedure for forming classes. In the third year, the NCTE project team randomly assigned teachers to rosters of students provided by the school.  

Table~\ref{tab:teacher_student_stats} provides key statistics about the transcripts, as well as teacher and student demographics. The majority of teachers in the data are white (65\%) and female (84\%). Whereas the student body is equally split in terms of gender, the majority are students of color (43\% African American, 23\% Hispanic/Latinx, 8\% Asian) and receive free or reduced lunch (67\%). This disparity between student-teacher demographics is in accordance with national statistics\footnote{\url{https://nces.ed.gov/}} \citep{Schaeffer2021}.

\begin{table*}[ht!]
\centering
\resizebox{.85\textwidth}{!}{%
\begin{tabular}{@{}ll@{}}
\toprule
\textbf{Variable}               & \textbf{Description}                                                                                                                                                                                     \\ \midrule
Turn-level annotations         & \begin{tabular}[c]{@{}l@{}}Annotations for on vs off task instruction, uptake of student \\ contributions, focusing questions and student reasoning. \end{tabular} \\  \midrule
Transcript-level observation scores & \begin{tabular}[c]{@{}l@{}}Observation scores by expert raters using two instruments:\\ CLASS \citep{pianta2008classroom} and MQI \citep{hill2008mathematical}.\end{tabular}                                                                          \\  \midrule
Student questionnaires            & \begin{tabular}[c]{@{}l@{}}Student survey responses about the classroom experience\\ and their household.\end{tabular}                                                                          \\  \midrule
Value-added scores                  & \begin{tabular}[c]{@{}l@{}}Teachers’ value added scores – i.e., an estimate of teachers’\\ contribution to students’ test performance.\end{tabular}                                             \\ \midrule
Student administrative data         & \begin{tabular}[c]{@{}l@{}}Administrative data on students, e.g. their test scores and \\ demographic information.\end{tabular}                                                                 \\ \midrule
Teacher questionnaires           & \begin{tabular}[c]{@{}l@{}}Teacher’s self reported information about their background,\\ beliefs and classroom practices.\end{tabular}                                                     \\ \bottomrule
\end{tabular}%
}
\caption{Variables linked to the NCTE transcript data. For full documentation, please refer to \citet{kane2015national}.}
\label{tab:ncte_variables}
\end{table*}

\subsection{Transcription \& Anonymization}

Lessons were captured by ThereNow using its Iris system,
\footnote{ThereNow is no longer in business, but their technology is now used by IrisConnect: \url{https://www.irisconnect.com/uk/products-and-services/video-technology-for-teachers/}}
which featured three cameras, a lapel microphone worn by teachers, and a bidirectional microphone for capturing student talk. The recordings were transcribed by professional transcribers working under contract to a commercial transcription company. Transcripts are fully anonymized: student and teacher names are replaced with terms like ``Student J'', ``Teacher'' or ``Mrs. H''. Inaudible talk, due to classroom noise and far field audio is transcribed as \emph{[Inaudible]}. If the transcriber was unsure of a particular word, they transcribed it within brackets, e.g. \emph{It is a city surrounded by [water]}. Square brackets are also used for other transcriber comments, such as \emph{[crosstalk]}, 
 and \emph{[laughter]}. Almost all teacher talk and the majority of student talk could be transcribed: only 4\% of teacher utterances and 21\% of student utterances contain an \emph{[Inaudible]} marker. Transcripts contain 5,733 words on average, 87.7\% of which are spoken by the teacher.

\subsection{Linked Variables}

The transcript data comes with a uniquely rich source of linked variables, summarized in Table~\ref{tab:ncte_variables}. These variables include turn-level annotations for various discourse features, classroom observation scores, demographic information about students and teachers, survey data and value-added scores. Please refer to  \citet{kane2015national} for a full documentation of these variables, except for the turn-level annotations, which we describe below.

\paragraph{Turn-level annotations.}

Table~\ref{tab:discourse-examples} includes examples for each discourse feature, annotated at the turn level. In prior work, experts annotated a sample of 2,348 utterance pairs --- exchanges between students and teachers --- for on vs off task instruction, teachers' uptake of student ideas \citep{demszky2021measuring} and focusing questions \citep{alic2022computationally}. The annotation process is described in its respective papers. We include the coding scheme on our Github along with the dataset.

In a separate annotation process, experts coded 2,000 student utterances for student reasoning, a key Common Core aligned student practice. The coding process was based on the MQI classroom observation item ``Student Provide Explanations'' \citep{hill2008mathematical}. To create a sample for labeling, we (i) hold out half of the transcripts for testing, (ii) from the remaining half, sample 30\% of transcripts from the top quartile in terms of their Student Provide Explanations MQI score, and 70\% from the rest, (iii) filter out student utterances shorter than 8 words, since they are unlikely to substantive reasoning, (iv) randomly sample up to 5 student utterances from each transcript, to balance representation across transcripts, (v) randomly sample 2,000 student utterances by assigning sampling weights proportionate to classroom diversity: combined percentage of African American and Hispanic/Latinx students in classroom. Each example was randomly assigned to one of two math coaches, who are also experts in the MQI coding instrument. Inter-rater agreement on calibration set (n=200) was 90\%.

\begin{table*}[]
\centering
\resizebox{.8\textwidth}{!}{%
\begin{tabular}{@{}ll@{}}
\toprule
\textbf{Discourse Feature} & \textbf{Example}                                                                                                         \\ \midrule
Student on Task   & S: We both have the same number of blue, and red, and yellow.                                                      \\ \midrule
Teacher on Task &
  \begin{tabular}[c]{@{}l@{}}T: Good, find the range.  Find the range.  Remember it's the span of the least to\\ the greatest number.\end{tabular} \\\midrule
Student Reasoning &
  \begin{tabular}[c]{@{}l@{}}S: Because if you add ninety-eight hundredths and five hundredths, I think it's going\\ to add up to, like, it's almost -- it's going to almost add up to a hundred.\end{tabular} \\ \midrule
High Uptake &
  \begin{tabular}[c]{@{}l@{}}S: I think these are Y axis and the X axis.\\ T: They do.  Sometimes they refer to them as X and Y axis.  It depends on the\\ type of graph. Okay.  You ready?\end{tabular} \\ \midrule
Focusing Question & \begin{tabular}[c]{@{}l@{}}S: Four fifths -- no, 80 percent.\\ T: How come you can't put it there?\end{tabular} \\ \bottomrule
\end{tabular}%
}
\caption{Examples for each discourse feature annotated at the turn-level. See \citet{demszky2021measuring} and \citet{alic2022computationally} for more details. }
\label{tab:discourse-examples}
\end{table*}

\section{Computational Analysis}

We illustrate how the NCTE transcript dataset can serve as a valuable resource for developing computational measures of classroom discourse. First, we train models that automatically identify dialogic discourse moves by leveraging the turn-level annotations. We then study how these discourse moves correlate with observation scores and teachers' value added scores.
\subsection{Pre-Processing Annotations}
We binarize annotations for each discourse feature in order to make the classification task consistent across features for the purposes of this paper. We also found that providing teachers with feedback using only their positive examples can be effective \citep{demszky2021can}, and thus in a first pilot experiment, one may not need to preserve fine-grained distinctions between negative, mediocre and positive examples.\footnote{Using these distinctions in teacher feedback is a promising direction of future work, given that contrasting examples can be an effective pedagogical tools \citep{schwartz2011practicing,sidney2015contrasting}. } 

Labels for on vs off task instruction and student reasoning are binary, so we simply consider the majority rater label for these discourse moves. Labels for uptake and focusing questions are on multi-level scales. We binarize them by assigning 1 as a label to examples where the majority of raters selected the top category (``high uptake", ``focusing question") and 0 to all others. 

\subsection{Supervised Classification}

We finetune RoBERTa \citep{liu2019roberta} on turn-level annotations for each discourse feature. We run finetuning for 5 epochs, a batch size of 8 x 2 gradient accumulation steps. The choice of this model and parameters were optimal for efficient iteration on a single TitanX GPU; we leave model exploration for future work. 

For Student on Task and Student Reasoning, the input to the model was a single student utterance. For Teacher on Task, the input to the model was a single teacher utterance. For High Uptake and Focusing Question, the model input was a student utterance and a subsequent teacher utterance, to match what annotators saw while labeling. We balance labels during training by oversampling the minority category to represent 50\% of labels, as we found this process yields better results.

\subsection{Regression Analysis}

Since our ultimate goal is to improve instruction via classroom discourse analysis, we need to understand if our NLP measures of discourse features indeed correlate with observation scores and student outcomes. To understand this question, we (i) follow the procedure described above to fine-tune classifiers for each discourse on \emph{all} annotations, (ii) predict discourse features for the entire NCTE transcript dataset, (iii) run regressions using classroom observation scores and value added scores as dependent variables.

\paragraph{Model.} We run a linear regression, clustering standard errors at the teacher level.  The models are captured by this equation:
\begin{equation}
    y_d = x_f\beta_1 + T\beta_2 + S\beta_3
+ \varepsilon
\end{equation}, 
where $y_d$ is a vector representing a dependent variable $d$, $x_f$ is a vector representing our predictions for a particular discourse feature $f$, $T$ is a matrix of teacher covariates, $S$ is a matrix of student covariates, $\beta_1,\beta_2,\beta_3$ are vectors of unknown parameters to be estimated and $\epsilon$ is a vector of residuals. 

As for discourse features $f$, we use discourse moves in Table~\ref{tab:discourse-examples} and also include baseline measures of student talk ratios (\% of student words and \% of student turns) for comparison. We estimate the effect of these features on six different dependent variables $d$, which include five items from observation scores and value added scores.

\paragraph{Observation scores as dependent variables.} To measure mathematics instructional quality, we use the main holistic item from the MQI instrument, a 5-level rating for lesson quality. The other four variables come from the CLASS scoring instrument: instructional dialogue, teacher sensitivity, teachers' regard for student perspectives and positive classroom climate. We picked these items \emph{a priori}---before conducting the regressions---based on their relevance to dialogic instruction. One could choose other items from these observation protocols to conduct similar analyses.  

Since observation scores are linked to transcripts, we first aggregate discourse move predictions to the lesson level. Specifically, we sum discourse feature predictions in each transcript, and divide them by the class duration (number of 7.5 minute segments in the transcript\footnote{We do not have consistent timestamps that are more granular than 7.5 minutes.}).

\paragraph{Value added scores as dependent variables.} Each teacher is linked to one value added score per year given that these are based on end-of-the-year standardized test scores. Therefore, we mean-aggregate lesson-level data obtained above to the teacher-year level when conducting regressions with value-added scores.

\paragraph{Covariates.} We include several covariates for teacher and classroom demographics. We include binary indicators for teacher gender and race/ethnicity and a numerical variable for years of experience. We include variables related to classroom composition in terms of student gender, race, free or reduced lunch status, special education status and limited English proficiency status -- see Table~\ref{tab:teacher_student_stats} for a list of variables.

\begin{table*}[ht!]
\centering
\resizebox{.5\textwidth}{!}{%
\begin{tabular}{@{}lcccc@{}}
\toprule
\textbf{Measure}      & \textbf{Accuracy} & \textbf{Precision} & \textbf{Recall} & \textbf{F1}    \\ \midrule
Student on Task   & 0.902    & 0.952     & 0.931  & 0.942 \\
Teacher on Task   & 0.867    & 0.932     & 0.914  & 0.923 \\
Teacher Uptake    & 0.768    & 0.719     & 0.674  & 0.688 \\
Focusing Question & 0.856    & 0.474     & 0.538  & 0.501 \\
Student Reasoning & 0.863    & 0.644     & 0.666  & 0.651 \\ \bottomrule
\end{tabular}%
}
\caption{Performance of RoBERTa on each discourse feature. Values are averages across a 5-fold cross validation.}
\label{tab:supervised}
\end{table*}
\section{Results}

\paragraph{Supervised classification.} Table~\ref{tab:supervised} shows the performance of our supervised classifiers, averaged across five-fold cross validation. The results show that we can train our model to automatically classify each discourse feature with moderate to high accuracy. The model performs best on classifying on vs off task instruction --- F1 score (harmonic mean of precision and recall) is $.942$ and $.923$ for student and teacher utterances, respectively. The model performs moderately well on higher-inference discourse moves, including Student Reasoning (F1 = $.651$), High Uptake (F1 = $.688$), and Focusing Questions (F1 = $.501$). We expect that model choice and hyperparameter tuning can improve the performance by 10-20\%. However, even the human raters are only able to reach moderate agreement on these measures \citep{demszky2021measuring,alic2022computationally} and other analogous ones in classroom observation instruments \citep{kelly2020using}. The moderate human agreement indicates that these measures are subjective, which may set an upper bound to the models' performance.

\paragraph{Correlation with outcomes.}

\begin{table*}[]
\centering
\renewcommand{\arraystretch}{1}
\resizebox{\textwidth}{!}{%
\begin{tabular}{lcccccc}
\hline
                         
 &
  \textbf{\begin{tabular}[c]{@{}c@{}}Value Added\\ Scores\end{tabular}} &
  \textbf{\begin{tabular}[c]{@{}c@{}}Math Instruction\\ Quality\end{tabular}} &
  \textbf{\begin{tabular}[c]{@{}c@{}}Instructional\\ Dialogue\end{tabular}} &
  \textbf{\begin{tabular}[c]{@{}c@{}}Teacher\\ Sensitivity\end{tabular}} &
  \textbf{\begin{tabular}[c]{@{}c@{}}Regard for Student\\ Perspectives\end{tabular}} &
  \textbf{\begin{tabular}[c]{@{}c@{}}Positive\\ Climate\end{tabular}}  \\ \cline{2-7} 
Student on Task   & 0.038+          & \textbf{0.022*}  & \textbf{0.032**} & \textbf{0.033**} & \textbf{0.024**} & \textbf{0.036**} \\
                  & (0.020)         & (0.010)          & (0.011)          & (0.008)          & (0.007)          & (0.007)          \\
Teacher on Task   & 0.038+          & \textbf{0.021*}  & \textbf{0.030**} & \textbf{0.034**} & \textbf{0.024**} & \textbf{0.035**} \\
                  & (0.020)         & (0.010)          & (0.010)          & (0.008)          & (0.007)          & (0.007)          \\
Teacher Uptake    & \textbf{0.121*} & \textbf{0.117**} & \textbf{0.083**} & \textbf{0.089**} & \textbf{0.058**} & \textbf{0.079**} \\
                  & (0.050)         & (0.032)          & (0.026)          & (0.019)          & (0.017)          & (0.017)          \\
Focusing Question & \textbf{0.234*}          & \textbf{0.233**} & \textbf{0.198**} & \textbf{0.132**} & \textbf{0.164**} & \textbf{0.115**} \\
                  & (0.104)         & (0.086)          & (0.072)          & (0.035)          & (0.044)          & (0.036)          \\
Student Reasoning & \textbf{0.191*}        & \textbf{0.313**} & \textbf{0.246**} & \textbf{0.144**} & \textbf{0.173**} & \textbf{0.120**} \\
                  & (0.091)         & (0.066)          & (0.050)          & (0.031)          & (0.035)          & (0.035)          \\
Student Turn \% & 1.044           & -0.047           & 0.718            & 0.214            & 0.125            & -0.172           \\
                  & (1.357)         & (0.528)          & (0.669)          & (0.574)          & (0.485)          & (0.560)          \\
Student Word \% & 0.359           & 0.721+           & \textbf{1.132*}  & 0.001            & 0.469            & 0.322            \\
                  & (0.792)         & (0.413)          & (0.541)          & (0.325)          & (0.395)          & (0.387)          \\
Observations      & 523            & 1557             & 1554             & 1554             & 1554             & 1554            \\ 
\hline
\end{tabular}%
}
\caption{The correlation of discourse features with outcomes, estimated at the transcript level. Each cell represents coefficients from a separate regression. Standard errors are enclosed in parentheses. Dependent variables are standardized. The ** $p < 0.01$, * $p < 0.05$, + $p < 0.1$}
\label{tab:outcomes}
\end{table*}

Table~\ref{tab:outcomes} shows the correlation of each discourse feature with outcomes. We also include two baseline discourse features that measure student talk ratios: the percentage of student turns and the percentage of student words in each transcript. We find that \textbf{all discourse features} predicted by our classifiers correlate significantly with each classroom observation item measuring instruction quality and classroom climate.

All dialogic discourse moves --- High Uptake, Focusing Questions and Student Reasoning --- also show a significant positive correlation with teachers' value added scores. Specifically, each additional High Uptake per 7.5 minute segment increases teachers' value added scores by 12\% of a standard deviation. Analogically, each additional instance of a Focusing Question and Student Reasoning per 7.5 minute segment increases teachers' value added scores by 23\% and 19\% of a standard deviation, respectively. Student on Task and Teacher on Task also correlate with marginal significance with value added scores.

Interestingly, baseline measures of student talk percentages do not in themselves correlate with observation scores or value added scores. One exception is student word percentage, which correlates positively with ratings for instructional dialogue. 
\section{Discussion}

We find that our measures of discourse moves, which we identified by consulting research on mathematics instruction, correlate with human raters' perceptions of lesson quality, and with students' learning outcomes. These results are significant in multiple ways.

\subsection{Implications \& Significance}

The fact that all of our measures correlate with MQI and CLASS observation scores indicate that the automated measures align with expert evaluations of instruction. This is a key result that provides external validation for these automated measures. The positive correlation of teacher value-added scores with measures for teacher uptake, focusing questions and student reasoning suggests that the use of these dialogic talk moves is associated with student learning. Substantively, this finding contributes evidence that classrooms where students are more deeply engaged with mathematical ideas --- and where teachers use their students' mathematical contributions --- are more likely to produce better achievement outcomes \citep{oconnor1993aligning,michaels2015conceptualizing}.

These findings become even more significant in the context of baseline measures of student engagement --- percentage of student turns and percentage of student words --- which do not show positive correlations with instruction quality and value added scores. These findings add to a collection of mixed results by related work, some of which show positive, some of which show no relationship between student talk time and learning outcomes \citep[see][for an overview]{sedova2019those}.

\subsection{Limitations}

\paragraph{Unmeasured covariates.}  A range of factors may affect instructional outcomes, only a subset of which could be measured with this data. Making strong claims about the link between discourse moves and instructional outcomes requires experimental validation. For example, the quality of the math task that the students are working may affect the discourse as well as learning outcomes. We can isolate the effect of discourse moves by randomly assigning teachers to learning opportunities that help them improve their use of these moves, and examining downstream impacts of these new talk moves on student outcomes. \citep{demszky2021can} has taken a similar approach successfully in an informal teaching context, but such a study is yet to be done in a K-12 context.

\paragraph{Generalizability.} Although the NCTE transcript dataset is the largest available dataset of U.S. classroom transcripts, it only captures a tiny fraction of U.S. classrooms and hence there are limitations to its representativeness. The data represents mostly white female teachers working in mid-size to large districts, so it would be valuable to collect new data from other types of districts and a more diverse teacher population. The fact that the data was collected a decade ago may pose limitations to its ongoing relevance; during the period under study (2010-2013), many schools were transitioning toward Common Core-aligned instruction in mathematics but yet lacked high-quality curriculum materials for doing so. That said, research in education reform has long attested to the fact that teaching practices have remained relatively constant over time \citep{cuban1993teachers,cohen2017reform} and that there are strong socio-cultural pressures that maintain the instructional status quo \citep{cohen1988teaching}. In general, it is important to carefully validate measures built on the NCTE data on a new domain to ensure that it is representative of the target population.

\paragraph{Limitations of linked data.} Education research has attested the limitations of standardized assessments in capturing student learning and reasoning \citep{sussman2019use}. Student questionnaires in the NCTE data can provide an alternative perspective on students' experiences and mathematics outcomes but these responses have a lot of missing values, and hence it may not provide robust estimates. Furthermore, understanding equity in instruction is a high priority for our research team and for the field more generally. However, studying equity within this data is challenging, since student speakers are not linked to administrative files containing student background and achievement variables. That said, such speaker-level demographic data is rarely available in instructional contexts, for important ethical reasons, and thus this limitation may encourage researchers to develop measures of instructional equity that leverage classroom-level, instead of speaker-level demographic information.

\subsection{Ethical Considerations}
We outline measures to safeguard students and teachers in this data and the users of it.

\paragraph{Consent \& privacy.} Both parents and teachers gave consent for the de-identified data to be retained and used in future research.
It is our highest priority that the identity of teachers and students in the data are kept private. Given that the none of the district and school names are disclosed, and that transcripts are fully de-identified, it is not possible to recover the identity of teachers and students.

\paragraph{Representation.} As \citet{madaio2022beyond} point out, applying AI in the educational domains comes with a risk of propagating and exaggerating existing inequities. As we describe in the paper, the data represents a largely low-income, demographically diverse student population. This means that the data can help with creating measures that are representative of low income students, students of color and students who are English language learners who receive the type of instruction captured in this dataset. As we point out above, the data should not be assumed to represent the diversity of identities and experiences of all students and teachers in U.S. classrooms and different forms of instruction. Furthermore, the data was annotated by raters whose demographics are largely representative of teacher demographics in the US\footnote{\url{https://nces.ed.gov/fastfacts/display.asp?id=28}} \citep{demszky2021measuring}, which, just like in this data, does not unfortunately match U.S. student demographics. Rater bias \cite{campbell2018observational} cannot be outruled especially given the subjectivity of these constructs.

\paragraph{Downstream application.} Users of this data have to agree to never use the data in a way that may cause harm to students and teachers, such as to build tools that discriminate against different groups of students and teachers or to surveil and punish teachers based on their practice. The sole purpose of this data should be to help us understand and facilitate student-centered and equitable instructional practice, and to empower historically marginalized teacher and student populations.

\subsection{Directions for Future Work}
The NCTE dataset opens up numerous directions for future work, some of which we are currently pursuing. First, one key direction is \textbf{building new NLP measures} of instruction. Observation protocols such as MQI, CLASS and the Culturally Responsive Instruction Observation Protocol (CRIOP) \citep{powell2016operationalizing} and related work by \citet{suresh2022talkmoves} and \citet{hunkins2022beautiful} can provide inspiration for various discourse features that can be measured in this data. Given the context-dependence of several instructional moves, new measures can incorporate more context beyond a single utterance or exchange between the student and the teacher. It would also be valuable to incorporate lesson-level metadata in the NLP model, e.g. lesson keywords, date, grade level, to create more context-specific measures. 

One can also conduct \textbf{bottom-up exploration of linguistic patterns} in the data to inform education research. For example, one could create an equity gap measure, leveraging academic outcomes and classroom demographics, and compare transcripts from classrooms with low equity gap with ones from classrooms with high equity gap. Doing so can help identify instructional correlates of equity gap, and help us understand how we can facilitate equitable instructional practices.

Third, since no two education settings are the same, it would be extremely valuable to \textbf{complement the NCTE dataset with other educational datasets} from a diverse range of settings, including various school subjects, informal and formal, online and in person, group and one-on-one settings, as well as data from multiple regions and countries and from multiple modalities, including text, audio and video. These datasets together can help us understand how effective instruction looks like across teaching contexts and modalities and build measures sensitive to these contextual differences.

Finally, one can \textbf{apply what we learn from this data to improve instruction}. Measures built on this data can power automated feedback tools \citep{suresh2021using,demszky2021can} and enable empirically-driven improvements to widely used professional development frameworks \citep[e.g.][NCTM guides\footnote{\url{https://www.nctm.org/pdguides/}}]{gregory2017my}. We hope that this resource can power efforts to address pressing issues in education, such as pandemic learning loss and equity gaps in mathematics instruction.

\section{Conclusion}

We introduce a dataset of elementary math classroom transcripts, associated with a rich source of linked variables. We train classifiers on turn-level annotations to predict discourse moves and show that predictions for these moves correlate significantly with observation scores and value added scores. These results demonstrate how the NCTE dataset can serve as a valuable resource for understanding classroom interactions and for powering tools that seek to facilitate instruction.

\section*{Acknowledgments}
We thank Shyamoli Sanghi for her assistance with finalizing the dataset. We also thank Jing Liu, Hannah Kleen, Zach Himmelsbach, Zid Mancenido amd Dan Jurafsky for the productive conversations and their collaboration. We are grateful to Dan McGinn for help with data sharing and to Christine Kuzdzal and Carol DeFreese for help with annotating student reasoning.

\bibliography{anthology,main}

\begin{thebibliography}{44}
\expandafter\ifx\csname natexlab\endcsname\relax\def\natexlab#1{#1}\fi

\bibitem[{Alexander(2008)}]{alexander2008dialogic}
{Robin John} Alexander. 2008.
\newblock \emph{Towards Dialogic Teaching: rethinking classroom talk (4th
  Edition)}.
\newblock Dialogos.

\bibitem[{Alic et~al.(2022)Alic, Demszky, Mancenido, Liu, Hill, and
  Jurafsky}]{alic2022computationally}
Sterling Alic, Dorottya Demszky, Zid Mancenido, Jing Liu, Heather Hill, and Dan
  Jurafsky. 2022.
\newblock Computationally identifying funneling and focusing questions in
  classroom discourse.
\newblock In \emph{Proceedings of the 17th Workshop on Innovative Use of NLP
  for Building Educational Applications (BEA 2022)}, pages 224--233.

\bibitem[{Bakhtin(1981)}]{bakhtin1986dialogic}
M.~M. Bakhtin. 1981.
\newblock \emph{The dialogic imagination: four essays}.
\newblock University of Texas Press.

\bibitem[{Braley and Murray(2018)}]{gapcorpus18}
McKenzie Braley and Gabriel Murray. 2018.
\newblock The group affect and performance (gap) corpus.
\newblock In \emph{Proceedings of the ICMI 2018 Workshop on Group Interaction
  Frontiers in Technology (GIFT)}.

\bibitem[{Brophy(1984)}]{brophy1984teacher}
Jere~E Brophy. 1984.
\newblock \emph{Teacher behavior and student achievement}.
\newblock 73. Institute for Research on Teaching, Michigan State University.

\bibitem[{Budzianowski et~al.(2018)Budzianowski, Wen, Tseng, Casanueva, Ultes,
  Ramadan, and Gasic}]{budzianowski2018multiwoz}
Pawe{\l} Budzianowski, Tsung-Hsien Wen, Bo-Hsiang Tseng, I{\~n}igo Casanueva,
  Stefan Ultes, Osman Ramadan, and Milica Gasic. 2018.
\newblock Multiwoz-a large-scale multi-domain wizard-of-oz dataset for
  task-oriented dialogue modelling.
\newblock In \emph{Proceedings of the 2018 Conference on Empirical Methods in
  Natural Language Processing}, pages 5016--5026.

\bibitem[{Campbell and Ronfeldt(2018)}]{campbell2018observational}
Shanyce~L Campbell and Matthew Ronfeldt. 2018.
\newblock Observational evaluation of teachers: Measuring more than we
  bargained for?
\newblock \emph{American Educational Research Journal}, 55(6):1233--1267.

\bibitem[{Cohen(1988)}]{cohen1988teaching}
David~K Cohen. 1988.
\newblock \emph{Teaching practice: Plus {\c{c}}a change}.
\newblock National Center for Research on Teacher Education East Lansing, MI.

\bibitem[{Cohen and Mehta(2017)}]{cohen2017reform}
David~K Cohen and Jal~D Mehta. 2017.
\newblock Why reform sometimes succeeds: Understanding the conditions that
  produce reforms that last.
\newblock \emph{American Educational Research Journal}, 54(4):644--690.

\bibitem[{Cuban(1993)}]{cuban1993teachers}
Larry Cuban. 1993.
\newblock \emph{How teachers taught: Constancy and change in American
  classrooms, 1890-1990}.
\newblock Teachers College Press.

\bibitem[{Demszky et~al.(2021{\natexlab{a}})Demszky, Liu, Hill, Jurafsky, and
  Piech}]{demszky2021can}
Dorottya Demszky, Jing Liu, Heather~C Hill, Dan Jurafsky, and Chris Piech.
  2021{\natexlab{a}}.
\newblock Can automated feedback improve teachers' uptake of student ideas?
  evidence from a randomized controlled trial in a large-scale online course.
  edworkingpaper no. 21-483.
\newblock \emph{Annenberg Institute for School Reform at Brown University}.

\bibitem[{Demszky et~al.(2021{\natexlab{b}})Demszky, Liu, Mancenido, Cohen,
  Hill, Jurafsky, and Hashimoto}]{demszky2021measuring}
Dorottya Demszky, Jing Liu, Zid Mancenido, Julie Cohen, Heather Hill, Dan
  Jurafsky, and Tatsunori~B Hashimoto. 2021{\natexlab{b}}.
\newblock Measuring conversational uptake: A case study on student-teacher
  interactions.
\newblock In \emph{Proceedings of the 59th Annual Meeting of the Association
  for Computational Linguistics and the 11th International Joint Conference on
  Natural Language Processing (Volume 1: Long Papers)}, pages 1638--1653.

\bibitem[{Devlin et~al.(2019)Devlin, Chang, Lee, and
  Toutanova}]{devlin2019bert}
Jacob Devlin, Ming-Wei Chang, Kenton Lee, and Kristina Toutanova. 2019.
\newblock Bert: Pre-training of deep bidirectional transformers for language
  understanding.
\newblock In \emph{Proceedings of the 2019 Conference of the North American
  Chapter of the Association for Computational Linguistics: Human Language
  Technologies, Volume 1 (Long and Short Papers)}, pages 4171--4186.

\bibitem[{Godfrey and Holliman(1997)}]{godfrey1997switchboard}
John~J Godfrey and Edward Holliman. 1997.
\newblock Switchboard-1 release 2.
\newblock \emph{Linguistic Data Consortium, Philadelphia}, 926:927.

\bibitem[{Gregory et~al.(2017)Gregory, Ruzek, Hafen, Mikami, Allen, and
  Pianta}]{gregory2017my}
A~Gregory, E~Ruzek, CA~Hafen, A~Yee Mikami, JP~Allen, and RC~Pianta. 2017.
\newblock My teaching partner-secondary: A video-based coaching model.
\newblock \emph{Theory into practice}, 56(1):38--45.

\bibitem[{Hill et~al.(2008)Hill, Blunk, Charalambous, Lewis, Phelps, Sleep, and
  Ball}]{hill2008mathematical}
Heather~C Hill, Merrie~L Blunk, Charalambos~Y Charalambous, Jennifer~M Lewis,
  Geoffrey~C Phelps, Laurie Sleep, and Deborah~Loewenberg Ball. 2008.
\newblock Mathematical knowledge for teaching and the mathematical quality of
  instruction: An exploratory study.
\newblock \emph{Cognition and instruction}, 26(4):430--511.

\bibitem[{Hunkins et~al.(2022)Hunkins, Kelly, and
  D'Mello}]{hunkins2022beautiful}
Nicholas Hunkins, Sean Kelly, and Sidney D'Mello. 2022.
\newblock “beautiful work, you're rock stars!”: Teacher analytics to
  uncover discourse that supports or undermines student motivation, identity,
  and belonging in classrooms.
\newblock In \emph{LAK22: 12th International Learning Analytics and Knowledge
  Conference}, pages 230--238.

\bibitem[{Jurafsky et~al.(1997)Jurafsky, Shriberg, and
  Biasca}]{Jurafsky:97-damsl}
Daniel Jurafsky, Elizabeth Shriberg, and Debra Biasca. 1997.
\newblock {S}witchboard {SWBD-DAMSL} {L}abeling {P}roject {C}oder's {M}anual,
  {D}raft 13.
\newblock Technical Report 97-02, University of Colorado Institute of Cognitive
  Science.

\bibitem[{Kane et~al.(2015)Kane, Hill, and Staiger}]{kane2015national}
T~Kane, H~Hill, and D~Staiger. 2015.
\newblock National center for teacher effectiveness main study. icpsr36095-v2.

\bibitem[{Kane et~al.(2013)Kane, McCaffrey, Miller, and Staiger}]{kane2013have}
Thomas~J Kane, Daniel~F McCaffrey, Trey Miller, and Douglas~O Staiger. 2013.
\newblock Have we identified effective teachers? validating measures of
  effective teaching using random assignment. research paper. met project.
\newblock \emph{Bill \& Melinda Gates Foundation}.

\bibitem[{Kelly et~al.(2020)Kelly, Bringe, Aucejo, and
  Fruehwirth}]{kelly2020using}
Sean Kelly, Robert Bringe, Esteban Aucejo, and Jane~Cooley Fruehwirth. 2020.
\newblock Using global observation protocols to inform research on teaching
  effectiveness and school improvement: Strengths and emerging limitations.
\newblock \emph{Education Policy Analysis Archives}, 28:62.

\bibitem[{Kelly et~al.(2018)Kelly, Olney, Donnelly, Nystrand, and
  D’Mello}]{kelly2018automatically}
Sean Kelly, Andrew~M Olney, Patrick Donnelly, Martin Nystrand, and Sidney~K
  D’Mello. 2018.
\newblock Automatically measuring question authenticity in real-world
  classrooms.
\newblock \emph{Educational Researcher}, 47(7):451--464.

\bibitem[{Liu and Cohen(2020)}]{liu2020measuring}
Jing Liu and Julie Cohen. 2020.
\newblock Measuring teaching practices at scale: A novel application of
  text-as-data methods.
\newblock \emph{EdWorking PaperNo}, pages 20--239.

\bibitem[{Liu et~al.(2019)Liu, Ott, Goyal, Du, Joshi, Chen, Levy, Lewis,
  Zettlemoyer, and Stoyanov}]{liu2019roberta}
Yinhan Liu, Myle Ott, Naman Goyal, Jingfei Du, Mandar Joshi, Danqi Chen, Omer
  Levy, Mike Lewis, Luke Zettlemoyer, and Veselin Stoyanov. 2019.
\newblock {RoBERTa: A robustly optimized BERT pretraining approach}.
\newblock \emph{arXiv preprint arXiv:1907.11692}.

\bibitem[{Madaio et~al.(2022)Madaio, Blodgett, Mayfield, and
  Dixon-Rom{\'a}n}]{madaio2022beyond}
Michael Madaio, Su~Lin Blodgett, Elijah Mayfield, and Ezekiel Dixon-Rom{\'a}n.
  2022.
\newblock Beyond “fairness”: Structural (in) justice lenses on ai for
  education.
\newblock In \emph{The Ethics of Artificial Intelligence in Education}, pages
  203--239. Routledge.

\bibitem[{Michaels and O’Connor(2015)}]{michaels2015conceptualizing}
Sarah Michaels and Catherine O’Connor. 2015.
\newblock Conceptualizing talk moves as tools: Professional development
  approaches for academically productive discussion.
\newblock \emph{Socializing intelligence through talk and dialogue}, 347:362.

\bibitem[{Michaels et~al.(2010)Michaels, O’Connor, Hall, and
  Resnick}]{michaels2010accountable}
Sarah Michaels, Mary~Catherine O’Connor, Megan~Williams Hall, and Lauren~B
  Resnick. 2010.
\newblock Accountable talk sourcebook: For classroom conversation that works.
\newblock \emph{Pittsburgh, PA: University of Pittsburgh Institute for
  Learning}.

\bibitem[{Nystrand et~al.(1997)Nystrand, Gamoran, Kachur, and
  Prendergast}]{nystrand1997opening}
Martin Nystrand, Adam Gamoran, Robert Kachur, and Catherine Prendergast. 1997.
\newblock \emph{Opening dialogue}.
\newblock New York: Teachers College Press.

\bibitem[{O'Connor and Michaels(1993)}]{oconnor1993aligning}
Mary~C O'Connor and Sarah Michaels. 1993.
\newblock Aligning academic task and participation status through revoicing:
  Analysis of a classroom discourse strategy.
\newblock \emph{Anthropology \& Education Quarterly}, 24(4):318--335.

\bibitem[{Pennebaker et~al.(2001)Pennebaker, Francis, and
  Booth}]{pennebaker2001linguistic}
James~W Pennebaker, Martha~E Francis, and Roger~J Booth. 2001.
\newblock Linguistic inquiry and word count: Liwc 2001.
\newblock \emph{Mahway: Lawrence Erlbaum Associates}, 71(2001):2001.

\bibitem[{Pianta et~al.(2008)Pianta, La~Paro, and Hamre}]{pianta2008classroom}
Robert~C Pianta, Karen~M La~Paro, and Bridget~K Hamre. 2008.
\newblock \emph{{Classroom Assessment Scoring System: Manual K-3.}}
\newblock Paul H Brookes Publishing.

\bibitem[{Powell et~al.(2016)Powell, Cantrell, Malo-Juvera, and
  Correll}]{powell2016operationalizing}
Rebecca Powell, Susan~Chambers Cantrell, Victor Malo-Juvera, and Pamela
  Correll. 2016.
\newblock Operationalizing culturally responsive instruction: Preliminary
  findings of criop research.
\newblock \emph{Teachers College Record}, 118(1):1--46.

\bibitem[{Reimer et~al.(2016)Reimer, Schenke, Nguyen, O'dowd, Domina, and
  Warschauer}]{reimer2016evaluating}
Lynn~C Reimer, Katerina Schenke, Tutrang Nguyen, Diane~K O'dowd, Thurston
  Domina, and Mark Warschauer. 2016.
\newblock Evaluating promising practices in undergraduate stem lecture courses.
\newblock \emph{RSF: The Russell Sage Foundation Journal of the Social
  Sciences}, 2(1):212--233.

\bibitem[{Rossi and Gnawali(2014)}]{coursera-iri2014}
Lorenzo~A. Rossi and Omprakash Gnawali. 2014.
\newblock {Language Independent Analysis and Classification of Discussion
  Threads in Coursera MOOC Forums}.
\newblock In \emph{Proceedings of the IEEE International Conference on
  Information Reuse and Integration (IRI 2014)}.

\bibitem[{Schaeffer(2021)}]{Schaeffer2021}
Katherine Schaeffer. 2021.
\newblock \href
  {https://www.pewresearch.org/fact-tank/2021/12/10/americas-public-school-teachers-are-far-less-racially-and-ethnically-diverse-than-their-students/}
  {America’s public school teachers are far less racially and ethnically
  diverse than their students}.

\bibitem[{Schwartz et~al.(2011)Schwartz, Chase, Oppezzo, and
  Chin}]{schwartz2011practicing}
Daniel~L Schwartz, Catherine~C Chase, Marily~A Oppezzo, and Doris~B Chin. 2011.
\newblock Practicing versus inventing with contrasting cases: The effects of
  telling first on learning and transfer.
\newblock \emph{Journal of educational psychology}, 103(4):759.

\bibitem[{Sedova et~al.(2019)Sedova, Sedlacek, Svaricek, Majcik, Navratilova,
  Drexlerova, Kychler, and Salamounova}]{sedova2019those}
Klara Sedova, Martin Sedlacek, Roman Svaricek, Martin Majcik, Jana Navratilova,
  Anna Drexlerova, Jakub Kychler, and Zuzana Salamounova. 2019.
\newblock Do those who talk more learn more? the relationship between student
  classroom talk and student achievement.
\newblock \emph{Learning and instruction}, 63:101217.

\bibitem[{Sidney et~al.(2015)Sidney, Hattikudur, and
  Alibali}]{sidney2015contrasting}
Pooja~G Sidney, Shanta Hattikudur, and Martha~W Alibali. 2015.
\newblock How do contrasting cases and self-explanation promote learning?
  evidence from fraction division.
\newblock \emph{Learning and Instruction}, 40:29--38.

\bibitem[{Stasaski et~al.(2020)Stasaski, Kao, and
  Hearst}]{stasaski-etal-2020-cima}
Katherine Stasaski, Kimberly Kao, and Marti~A. Hearst. 2020.
\newblock \href {https://doi.org/10.18653/v1/2020.bea-1.5} {{CIMA}: A large
  open access dialogue dataset for tutoring}.
\newblock In \emph{Proceedings of the Fifteenth Workshop on Innovative Use of
  NLP for Building Educational Applications}, pages 52--64, Seattle, WA, USA
  → Online. Association for Computational Linguistics.

\bibitem[{Suresh et~al.(2021)Suresh, Jacobs, Clevenger, Lai, Tan, Martin, and
  Sumner}]{suresh2021using}
Abhijit Suresh, Jennifer Jacobs, Charis Clevenger, Vivian Lai, Chenhao Tan,
  James~H Martin, and Tamara Sumner. 2021.
\newblock Using ai to promote equitable classroom discussions: The talkmoves
  application.
\newblock In \emph{International Conference on Artificial Intelligence in
  Education}, pages 344--348. Springer.

\bibitem[{Suresh et~al.(2022)Suresh, Jacobs, Harty, Perkoff, Martin, and
  Sumner}]{suresh2022talkmoves}
Abhijit Suresh, Jennifer Jacobs, Charis Harty, Margaret Perkoff, James~H
  Martin, and Tamara Sumner. 2022.
\newblock The talkmoves dataset: K-12 mathematics lesson transcripts annotated
  for teacher and student discursive moves.
\newblock \emph{arXiv preprint arXiv:2204.09652}.

\bibitem[{Suresh et~al.(2019)Suresh, Sumner, Jacobs, Foland, and
  Ward}]{suresh2019automating}
Abhijit Suresh, Tamara Sumner, Jennifer Jacobs, Bill Foland, and Wayne Ward.
  2019.
\newblock Automating analysis and feedback to improve mathematics teachers’
  classroom discourse.
\newblock In \emph{Proceedings of the AAAI conference on artificial
  intelligence}, volume~33, pages 9721--9728.

\bibitem[{Sussman and Wilson(2019)}]{sussman2019use}
Joshua Sussman and Mark~R Wilson. 2019.
\newblock The use and validity of standardized achievement tests for evaluating
  new curricular interventions in mathematics and science.
\newblock \emph{American Journal of Evaluation}, 40(2):190--213.

\bibitem[{Wells(1999)}]{wells1999dialogic}
Gordon Wells. 1999.
\newblock \emph{Dialogic inquiry: Towards a socio-cultural practice and theory
  of education}.
\newblock Cambridge University Press.

\end{thebibliography}

\clearpage
\appendix

\end{document}